\pdfoutput=1

\documentclass[11pt]{article}

\usepackage[final]{acl}

\usepackage{times}
\usepackage{latexsym}

\usepackage[T1]{fontenc}

\usepackage[utf8]{inputenc}

\usepackage{microtype}

\usepackage{inconsolata}

\usepackage{graphicx}
\usepackage{inconsolata}
\usepackage{bm}
\usepackage{amsmath}
\usepackage{amsfonts}
\usepackage{xspace}
\usepackage{multicol}
\usepackage{multirow}
\usepackage{subfigure}
\usepackage{makecell}
\usepackage{listings}
\usepackage{adjustbox}
\usepackage{booktabs}
\usepackage{hyperref}
\usepackage{pifont}
\usepackage{enumitem}
\usepackage{amsthm}

\usepackage[capitalize,noabbrev]{cleveref}

\newtheorem{theorem}{Theorem}[section]

\newtheorem{lemma}[theorem]{Lemma}

\newtheorem{definition}[theorem]{Definition}
\newtheorem{assumption}[theorem]{Assumption}

\newtheorem{obs}{Observation}[section]

\newcommand{\esp}[1]{\mathbb{E}\left[#1\right]}
\newcommand{\espk}[1]{\mathbb{E}_{k-1}\left[#1\right]}

\usepackage[textsize=tiny]{todonotes}

\title{CoMMIT: Coordinated Multimodal Instruction Tuning}

\author{%
Xintong Li$^{1}\thanks{These authors contributed equally to this work.}$ \quad 
Junda Wu$^{1*}$ \quad 
Tong Yu$^{2}$ \quad 
Yu Wang$^1$ \quad 
Xiang Chen$^2$ \quad \\
\textbf{Jiuxiang Gu}$^2$ \quad
\textbf{Lina Yao}$^{3,4}$ \quad  
\textbf{Julian McAuley}$^{1}$ \quad
\textbf{Jingbo Shang}$^{1}$ \quad \\ 
$^1$University of California, San Diego \quad 
$^2$Adobe Research \quad \\
$^3$The University of New South Wales \quad
$^4$CSIRO's Data61  \\
\texttt{\{xil240,juw069,yuw164,jmcauley,jshang\}@ucsd.edu} \\
\texttt{\{tyu,xiangche,jigu\}@adobe.com} \quad
\texttt{lina.yao@data61.csiro.au} \\
}

\begin{document}
\maketitle

\begin{abstract}
Instruction tuning in multimodal large language models (MLLMs) generally involves cooperative learning between a backbone LLM and a feature encoder of non-text input modalities.
The major challenge is how to efficiently find the synergy between the two modules so that LLMs can adapt their reasoning abilities to downstream tasks
while feature encoders can adjust to provide more task-specific information about its modality. 
In this paper, we analyze the MLLM instruction tuning from both theoretical and empirical perspectives, 
where we find the unbalanced learning between the feature encoder and the LLM can cause problems of oscillation and biased learning that lead to sub-optimal convergence.
Inspired by our findings, we propose a Multimodal Balance Coefficient that enables quantitative measurement of the balance of learning.
Based on this, we further design a dynamic learning scheduler that better coordinates the learning between the LLM and feature encoder, alleviating the problems of oscillation and biased learning. 
In addition, we introduce an auxiliary regularization on the gradient to promote updating with larger step sizes, 
which potentially allows for a more accurate estimation of the proposed MultiModal Balance Coefficient and further improves the training sufficiency. 
Our proposed approach is agnostic to the architecture of LLM and feature encoder, so it can be generically integrated with various MLLMs.
We conduct experiments on multiple downstream tasks with various MLLMs, demonstrating the proposed method is more effective than the baselines in MLLM instruction tuning.
\end{abstract}

\section{Introduction} \label{sec:intro}

\begin{figure}[t]
    \centering
    \subfigure[]{
        \centering
        \includegraphics[width=0.22\textwidth, page=2, clip, viewport=20 30 120 125]{figures/fig1.pdf}
        \label{fig:demo-2}
    }
    \subfigure[]{
        \centering
        \includegraphics[width=0.20\textwidth, page=1, clip, viewport=20 20 120 125]{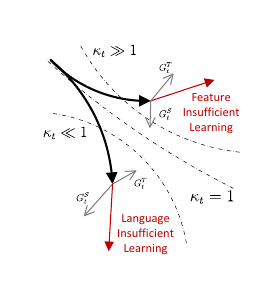}
        \label{fig:demo-1}
    }
    \caption{
        Illustration of (a) the oscillation problem and (b) the biased learning problem, caused by imbalanced multimodal learning.
        The optimization trajectories are shown in solid bold lines and the multimodal gradients at the current step $t$ are in solid thin lines.
    }
    \vspace{-1em}
    \label{fig:demo}
\end{figure}

Multimodal instruction tuning aligns pre-trained multimodal large language models (MLLMs) with specific downstream tasks by fine-tuning MLLMs to follow arbitrary instructions \citep{dai2024instructblip,zhang2023video,zhao2024slit,lu2023empirical,han2023imagebind,wu2024futga,wang2024instructgraph,wu2025doc,wu2025pdb}.
Leading pre-trained Multimodal Large Language Models (MLLMs) typically share similar architectures \citep{li2023blip,liu2024visual,tang2023salmonn,chu2023qwen}.
Specifically, the non-text data (image, audio, etc) is first encoded by a feature encoder into embedding tokens.
Then, these encoded embeddings are inserted into language prompts,
creating multimodal sequence inputs for the LLMs.
Effective multimodal understanding and reasoning in MLLMs depend on the model's ability to learn aligned multimodal features using its feature encoder (\emph{e.g.}, \cite{li2023blip}), 
and on leveraging the pre-trained capabilities of its backbone LLM (\emph{e.g.}, \cite{touvron2023llama, chiang2023vicuna}) to interpret these multimodal inputs.
This generally involves a two-prolonged learning process: (1) \textbf{LLM Adaptation.} Encoded non-text features (\emph{e.g.}, visual and auditory) in downstream tasks may not be perfectly aligned with pre-trained text features, thus requiring the backbone LLM to adapt its pre-trained parameters to recognize these new, non-text modality tokens. 
(2) \textbf{Feature Encoder Adaptation.} While LLMs possess strong reasoning ability from their pre-trained, it requires the feature encoders to be fine-tuned to extract task-specific information for evidence of reasoning.
Cooperative balancing of these two learning stages is crucial for effective instruction tuning of MLLMs.
When the learning is biased on the LLMs with (1), the insufficiently learned feature encoder can lead to information loss \citep{bai2024hallucination,tong2024eyes,wu2025mitigating}, hindering the LLM's ability to reason effectively due to a lack of adequate evidence from non-text modalities.
Conversely, if the learning is biased on the feature encoder with (2), the LLMs can be insufficiently adapted and struggle to interpret non-text modalities. As a result, it will cause the hallucination problem \citep{bai2024hallucination,rawte2024visual,wu2024personalized,wu2024visual} due to the strong language prior inherent in the backbone LLMs. 
Therefore, it is essential to balance the learning between the feature encoder and backbone LLM, so that the learning is not overly biased on either of the two modules.

In this paper, we first propose a multimodal balance coefficient that quantifies the learning balance between the feature encoder and the backbone LLM in MLLM instruction tuning.
Based on theoretical analysis and empirical observations, 
we identify two types of learning dilemmas that can be quantitatively measured by our proposed multimodal  balance coefficient:
\textit{i)} the oscillation  problem and \textit{ii)} the biased learning problem, as illustrated in Figure \ref{fig:demo}.
Specifically, Figure \ref{fig:demo-2} demonstrates the oscillation problem where the learning is \textit{alternatively} favoring either the feature encoder or the LLM.
This oscillation impedes the convergence of optimization and undermines learning efficiency since the learning is hardly progressing in consistent directions.
On the other hand, Figure \ref{fig:demo-1} shows the biased learning problem where the training \textit{consistently} favors either the LLM or the feature encoder.
In such cases, the gradient descent primarily only updates either the LLM or the feature encoder, resulting in insufficient learning of the other module.
This diminishes the effectiveness of gradient descent since the under-trained module (LLM or feature encoder) will not be capable of contributing sufficient information to the generation outputs.

To address these challenges, we propose \textbf{Co}ordinated \textbf{M}ulti\textbf{M}odal \textbf{I}nstruction \textbf{T}uning (\textbf{CoMMIT}), which regularizes the training with a coordinated learning rate scheduler (Section \ref{method}).
This scheduler dynamically adjusts the learning rates of the feature encoder and LLM according to the proposed multimodal balance coefficient,
ensuring sufficient gradient descent for both the feature encoder and LLM while mitigating the oscillation problem.
We also introduce a regularization loss that promotes larger update steps during training, further alleviating gradient diminishing.
We theoretically analyze the convergence rate and demonstrate that we can achieve accelerated convergence when optimizing with \textbf{CoMMIT} (Section \ref{sec:theory}).
We summarize our main contributions as follows:
\begin{itemize}
    \item We introduce a theoretical framework to uncover the pitfalls of the learning imbalance problem in MLLM instruction tuning, which can cause MLLM insufficient learning and the oscillation problem.
    \item Based on the theoretical analysis and empirical observation, we propose \textbf{CoMMIT} to balance multimodal learning progress by dynamically coordinating learning rates on the feature encoder and LLM.
    \textbf{CoMMIT} also enforces a gradient regularization that encourages larger step sizes and improves training efficiency.
    \item Applying \textbf{CoMMIT} introduces a novel term in the convergence rate analysis. Theoretical analysis proves that this term is always greater than one, leading to faster convergence. 
    We also demonstrate that the theorem can be generalized across various optimizers. 
    \item Empirical results on downstream tasks in vision and audio modalities with various LLM backbones show the efficiency and effectiveness of the proposed methods. 
    We demonstrate that \textbf{CoMMIT} can better coordinate multimodal learning progress with balanced training between the feature encoder and the LLM. 
\end{itemize}

\section{Related Works} 
MLLMs have become a new paradigm to empower multimodal learning with advanced language reasoning capabilities, 
such as with vision \citep{li2023blip,liu2024visual,wang2024visionllm,maaz2023video,zhang2023video,huang2023vtimellm,yan2024list}, and audio \citep{huang2023audiogpt, tang2023salmonn,gardner2023llark}.
Despite good generalizability and zero-shot performance of existing large language models (LLMs),
the discrepancy between different modalities can be one of the greatest challenges for LMMs to achieve comparable reasoning performance as LLMs.
To bridge the multimodality gap and align with downstream tasks, several works focus on two-fold considerations: feature (modality) alignment and reasoning alignment.
The most common approach for feature alignment is to encode the source modality feature to semantic tokens within the LLMs' embedding feature space.
By adding the modality-specific tokens \citep{wang2024tuning,liu2021unified,zhang2024vpgtrans} as soft prompt inputs \citep{liu2021p, xie2023few, wu2023few, wu2024infoprompt}, the backbone LLMs can process these tokens with language tokens as a unified sequence.
However, the newly added semantic tokens cannot be understood by LLMs directly for language reasoning, due to the limited text-only pretraining of LLMs.
Such misalignment problems will lead to textual hallucination problems, namely \textit{linguistic bias} \citep{ko2023large,tang2023video}, in which the language models reason only based on their language prior.
Thus, multimodal alignment should be achieved by additional adaptation of the LLM itself with multimodal instruction tuning.

\section{Preliminaries} \label{sec:pre}

Given a pair of non-text input $I^S$ (images, audio, etc) and instruction prompt $I^X$ of nature language, the instruction-tuned MLLM should comprehend the semantics of $I^S$ and generate outputs that comply with the instruction specified in $I^X$.
State-of-the-art MLLM instruction tuning generally adopts similar diagrams of training \citep{gardner2023llark,li2023blip,liu2024visual}, which involves cooperative training between a feature encoder $S$ and a pretrained LLM $X$. 
Specifically, $S$ first encodes the multimedia input $I^S$ into the embedding space of $X$. 
Then, the encoded $I^S$ is inserted into 
 the instruction prompt $I^X$ as input that conditions the output generation of $X$,
\begin{equation}
    P_{S, X}(\hat{y}_k|I^S,I^X,y_{j<k}) = X(\hat{y}_k|S(I^S),I^X,y_{j<k}),
\end{equation}

\noindent where $\hat{y}_{k}$ is the $k$th predicted token and $y_{j<k}$ denotes the first $k-1$ of the expected ground truth tokens in auto-regressive generation, $k=1, \cdots, K$.
The training loss is the cross-entropy defined by the predicted distribution 
 on $\hat{y}_k $ and the $k$th ground truth token $y_k$\citep{liu2024visual,ouyang2022training},
\begin{equation} \label{eq:xe}
\begin{split}
    L(&Y\!=\!\{y_k\}_{k=1}^K\,\,|\,\,I_S, I^T) = \\
    &-\frac{1}{K}\sum_{k=1}^{K} y_k \log  P_{S,X}(\hat{y}_k|I^S,I^X,y_{j<k}),
\end{split}
\end{equation}
The learning objective is to find the optimal $X$ and $S$ by minimizing the loss function. As mentioned in Section \ref{sec:intro}, the learning can either be inefficient by oscillating between the optimization of the two modules or insufficient by biasing on one of $S$ and $X$. 
Our goal is to find a balance between the learning of $X$ and $S$, so to accelerate the convergence while ensuring that both $S$ and $X$ are sufficiently trained.

\section{Measurement of Learning Balance in MLLM Instruction Tuning}\label{sec:balance}

To assess the balance between the updates on $X$ and $S$, we first measure the significance of each update separately with $X$ and $S$. Formally, for the $t$-th step of training, we define $d(P_{X_t,S_t}||P_{X_{t+1},S_t})$ and $d(P_{X_t,S_t}||P_{X_{t},S_{t+1}})$ that  quantify the significance of updates on $X$ and $S$, respectively, by measuring the shift in output distributions, 

\begingroup     %
  \small
  \begin{align}
    d&(P_{X_t,S_t}||P_{X_{t+1},S_t}) = \label{eq:gs}\\  
    &\frac{1}{K}\sum_k\mathbb{KL}\left(P_{S_t,X_t}(\hat{y}_k|I^S, I^X)||P_{S_{t+1},X_t}(\hat{y}_k|I^S, I^X)\right),  \notag\\
    d&(P_{X_t,S_t}||P_{X_{t},S_{t+1}}) = \label{eq:gx}\\
    &\frac{1}{K}\sum_k\mathbb{KL}\left(P_{S_t,X_t}(\hat{y}_k|I^S, I^X)||P_{S_{t},X_{t+1}}(\hat{y}_k|I^S, I^X)\right) \notag
  \end{align}
\endgroup

where we use the subscript $t$ to index the trained steps and $\mathbb {KL}(\cdot||\cdot)$ is the KL divergence. Based on \eqref{eq:gs} and \eqref{eq:gx}, we define the Multimodal Balance Coefficient that measures the balance between training on $X$ and $S$.

\begin{definition}[Multimodal balance coefficient]
 For time step $t$ of joint training on $X$ and $S$,
the Multimodal Balance Coefficient $\kappa_t$ is measured with the respective learning steps on the feature encoder $S_t$ and the LLM $X_t$.
\begin{equation}\label{eq:kappa}
    \kappa_t = \frac{d(P_{X_t,S_t}||P_{X_{t+1},S_t})}{d(P_{X_t,S_t}||P_{X_{t},S_{t+1}})}.
\end{equation}
\end{definition}
$\kappa_t$ with a value always near 1 indicates that the learning is balanced between the feature encoder and the LLM.
On the contrary, $\kappa_t>>1$ and $\kappa_t\rightarrow 0$ 
suggests that the learning biased by leaning toward $X$ and $S$, respectively. $\kappa_t$ with high variance corresponds to learning that oscillates between optimizing on $X$ or $S$. To further illustrate this, we derive Theorem 1 which estimated the gradient on $X$ and $S$ during training.

\begin{theorem} \label{thm1}
Let $G^X_t$  and $G^S_t$ be the gradient on $X$ and $S$ at time step $t$, we can derive the multimodal gradient estimated bounds,

\begin{equation}\label{eq:gd}
    \|G^X_t\| \leq  (\kappa_t + 1) H^S_t, \quad 
    \|G^S_t\| \leq (\frac{1}{\kappa_t} + 1) H^X_t,
\end{equation}
    where $H^S_t$ and $H^T_t$ represent the individual learning steps of $S$ and $X$, respectively. These are given by,
\begin{align}\label{eq:H}
    H^S_t &= \left(\|I^{X}_t\| + \|S_t(I^S_t)\|\right)^{-1} \|logits(P_{X_{t},S_{t+1}})\|, \nonumber \\
    H^X_t &= \left\|I^{S}_t\right\|^{-1} \|logits (P_{X_{t+1},S_{t}})\|.
\end{align}

\end{theorem}
The detailed proof is provided in Appendix~\ref{sec:lrbalance}.

Within the metric space of probability distribution $(P_{S,X},d)$, the values of $H^S_t$ and $H^T_t$ are bounded by a finite norm.
$H^S_t$ and $H^T_t$ are also lower-bounded, assuming multimodal gradients are not diminishing which we alleviate by proposing a regularization on the gradient in Section \ref{method}. 
Therefore, $\kappa_t$ can account the most for the gradient upper bound in \eqref{eq:gd} so it is suitable to measure the learning imbalance problem.

\section{Empirical Observations of Learning Dilemmas in MLLM Instruction Tuning}
\label{sec:emp}
In this section, we illustrate the learning dilemmas described in Section \ref{sec:intro} by observing the dynamics of $\kappa_t$ is different experiment settings.
The experiment is conducted by fine-tuning the  BLIP-2 \citep{li2023blip} model on 
 TextVQA\citep{singh2019towards}, a widely used dataset for visual question question answering \citep{dai2024instructblip,yin2024lamm}.  %
To probe the problems of oscillation and learning bias, we consider the following three learning strategies:
(1) \textbf{Synced LR} is trained by setting the learning rate of both $X$ and $S$ to $1e-4$;
(2) \textbf{Language LR $\uparrow$} increases learning rate of $X$ to $1e-3$;
(2) \textbf{Encoder LR $\uparrow$} increases the  learning rate of $S$ to $1e-3$.

\subsection{The Dilemma of Oscillation}\label{sec:pre-imbalance}
\begin{figure*}[ht]
    \centering
    \includegraphics[width=.95\linewidth]{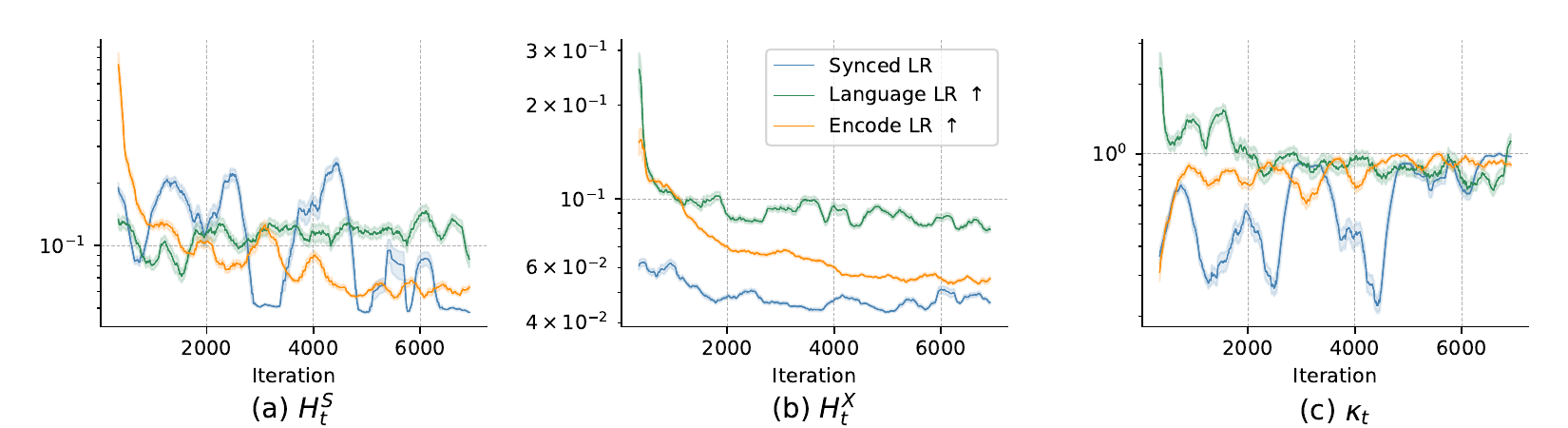}
    \caption{Learning curves of the variables $H^S_t$, $H^X_t$, and $\kappa_t$ to measure learning balance in multimodal instruction tuning.} %
    \label{fig:pre-1}
    \vspace{-3mm}
\end{figure*}

To quantitatively understand the oscillation problem in MLLM instruction tuning,
we show the learning curves (Figure \ref{fig:pre-1}) of the measurement variables $H^S_t$, $H^X_t$, and $\kappa_t$ proposed in \eqref{eq:gd}.

\begin{obs} 
\label{obs:lr_0}
As shown in Figure \ref{fig:pre-1}(c), the multimodal learning process can suffer from significant oscillation problems with highly variant $\kappa_t$
in the \textbf{Synced LR} setting.
\end{obs}

Specifically, the learning curve of $\kappa_t$ in the \textbf{Synced LR} setting exhibits high variance near the value of $1$ 
, which is a showcase of the oscillation problems that signify training instability.
This will cause inefficient training since the learning is alternating between $X$ and $S$, instead of progressing in consistent directions.
We demonstrate such inefficiency in Figures \ref{fig:exp-loss-1} and \ref{fig:exp-loss-2}, showing that its convergence is slower compared to our proposed approach, which balances the learning process with a more stable $\kappa_t$.
Further, it is interesting to find in Figure \ref{fig:pre-1} that the feature encoder $S$ is more unstable than the language model $X$, by comparing $H_t^S$ in Figure \ref{fig:pre-1}(a) and $H_t^X$ in Figure \ref{fig:pre-1}(b).
By increasing the learning rate either $X$ (\textbf{Encoder LR}) or $S$ (\textbf{Language LR}), we can observe that the three metrics in Figure \ref{fig:pre-1} are stabilized. In the next section, we show that such stabilization is at the expense of biased learning that causes insufficient training on $X$ or $S$.

\subsection{The Dilemma of Biased Learning}\label{sec:pre-insufficient}

\begin{figure*}[ht]
    \centering
    \includegraphics[width=.95\linewidth]{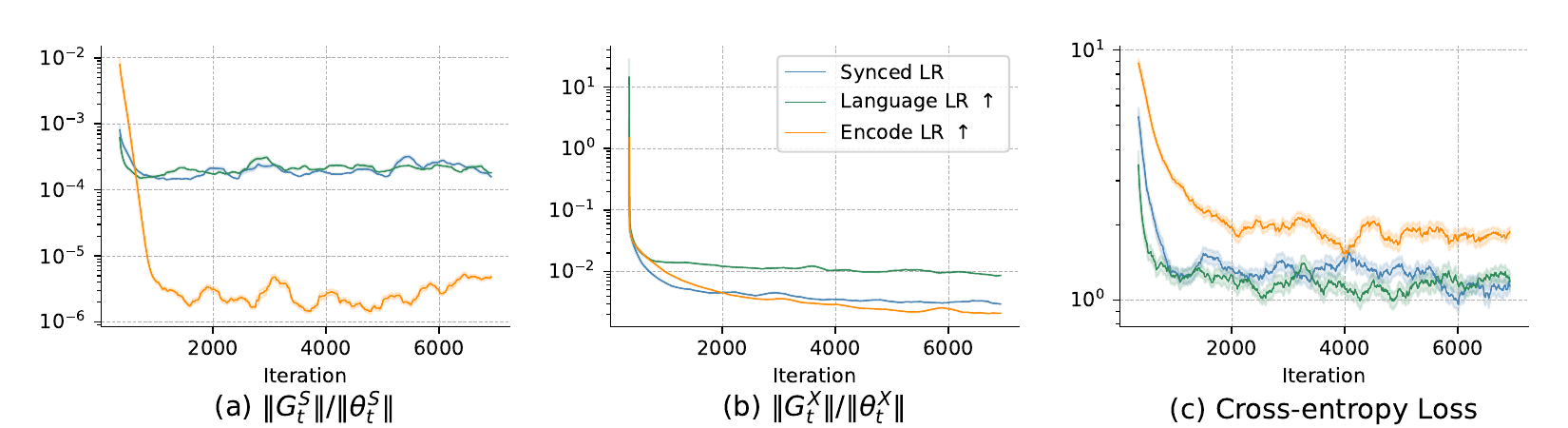}
    \caption{
    The learning curves of normalized learning gradient $\|G^S_t\|/\|\theta_{t}^S\|$ and $\|G^X_t\|/\|\theta_{t}^X\|$ for the feature encoder and language model respectively, as well as the cross-entropy training losses.}
    \label{fig:pre-2}
    \vspace{-3mm}
\end{figure*}

The \textbf{Encoder LR $\uparrow$} and \textbf{Language LR $\uparrow$} in Figure \ref{fig:pre-1} demonstrated the biased learning problem, where the training is biased on either $S$ or $X$.
Let $\theta^S_t$ and $\theta^X_t$ be the parameters of $S$ and $X$ at time step $t$.
We further show three metrics with the same backbone MLLM and training data as in Section \ref{sec:pre-imbalance}: 
(1) the normalized learning gradient $\|G^S_t\|/\|\theta_{t}^S\|$ of the feature encoder $S$ in Figure \ref{fig:pre-2}(a), 
(2) the normalized learning gradient $\|G^X_t\|/\|\theta_{t}^X\|$ of the language model $X$ in Figure \ref{fig:pre-2}(b),
and (3) the cross-entropy loss in Figure \ref{fig:pre-2}(c).
These metrics help understand the impact of biased learning on either $X$ or $S$  in MLLM instruction tuning.

\begin{obs} 
\label{obs:lr_1}
In Figure \ref{fig:pre-2}(c), \textbf{Encoder LR $\uparrow$} and  \textbf{language LR $\uparrow$} that are biased on either $X$ or $S$  can slow the convergence of the MLLM with gradient diminishing and inferior training performance.
\end{obs}

Such diminishing gradient would result in insufficient training on $X$ or $S$ with gradient descent.
For example, we can observe in Figure \ref{fig:pre-2}(a) and Figure \ref{fig:pre-2}(b) that the \textbf{Encoder LR} can simultaneously cause the gradient diminishing in both $X$ and $S$, with the cross-entropy converging to a higher value in Figure \ref{fig:pre-2}(c).
In such cases, it is necessary to strategically balance the learning between different modules, so the training is not leaning toward either $X$ or $S$.
In Figure \ref{fig:exp-kappa}, we show that our proposed approach can reduce such bias while not inducing oscillation, \emph{i.e.}, by learning with less variant $\kappa_t$ that is valued close to 1.

\section{CoMMIT: Coordinated Multimodal Instruction Tuning} \label{method}

Based on the observations in Section \ref{sec:emp}, the learning rate on $X$ should be boosted when the training is leaning toward $S$ ($\kappa_t\rightarrow 0$ ), and vice versa. 
So motivated, we propose a dynamic learning rate scheduling method to coordinate multimodal learning between $X$ and $S$,
which alleviates the oscillation problems while ensuring the training is not biased on either $X$ or $S$.

Inspired by damping strategies in optimization \citep{lucas2018aggregated,tanaka2021noether,wei2021stochastic}, 
we use the proposed learning balance metric $\kappa_t$ in \eqref{eq:kappa} as the damping parameter that facilitates balanced multimodal learning.
Specifically,
we track the $N_{\kappa}$ moving average of $\kappa_t$ through the learning process,
\begin{equation}\label{eq:dyn-lr}
    \Tilde{\kappa}_t = \frac{1}{N_{\kappa}}\sum_{i=1}^{N_{\kappa}} \kappa_{t-i+1},
\end{equation}
then dynamically adjust  learning rates of $X$ and $S$ in accordance.
Let $\beta^X_t$ and  $\beta^S_t$ be the learning rates on $X$ and $S$ at time step $t$. We adjust the learning rates by,
\begin{equation}
    \beta_t^T=\frac{2\alpha}{\Tilde{\kappa}_t + 1}, \,\,\beta_t^S=\frac{2\alpha}{ 1/\Tilde{\kappa}_t + 1},
\end{equation}
where $\alpha$ is the base learning rate.

During training, the diminishing $H^S_t$ and $H^T_t$ as observed in Figure \ref{fig:pre-2} can cause higher estimation errors in $\Tilde{\kappa}_t$.
To address these,
we propose an auxiliary regularization that encourages large step sizes for both $X$ and $S$, which mitigates gradient diminishing. 
Specifically, we want to encourage larger distribution drifts $d(P_{X_t,S_t}||P_{X_{t+1},S_t})$ for $X$ and $d(P_{X_t,S_t}||P_{X_{t},S_{t+1}})$ for $S$, apart from gradient descending on the cross-entropy loss in \eqref{eq:xe}. 
The gradient update for our proposed CoMMIT at time step $t$ is, 
\begin{align}
    \label{eq:fea-weight}
    \theta_{t+1}^X \leftarrow \theta_{t}^X  & - 
    \beta_t^X \cdot \nabla_{\theta^X} L(S_t(\Tilde{X}_{\theta_{t}^X})) \\ \nonumber
    & + \beta_t^X \cdot \nabla_{\theta^X} d(P_{X_t,S_t}||P_{X_{t+1},S_t}), \\
    \label{eq:lm-weight}
    \theta_{t+1}^S \leftarrow \theta_{t}^S & - 
    \beta_t^S \cdot \nabla_{\theta^S} L(T_t(S_t(X));\theta_{t}^S) \\ \nonumber
    & + \beta_t^S \cdot \nabla_{\theta^S} d(P_{X_t,S_t}||P_{X_{t},S_{t+1}}).
\end{align}
Note that the distribution drifts $d(P_{X_t,S_t}||P_{X_{t+1},S_t})$ and $d(P_{X_t,S_t}||P_{X_{t},S_{t+1}})$ does not involve ground truth labels. 

\paragraph{Theoretical Convergence Analysis.}
We further provide a theoretical analysis of our proposed CoMMIT framework in Appendix~\ref{sec:theory}. 
Specifically, we establish a new convergence bound under sufficient assumptions for the coordinated learning rate and regularization strategy, 
demonstrating that the proposed method can achieve a provably faster convergence rate compared to standard imbalanced instruction tuning approaches.
We further provide the detailed proof of the proposed theorems in Appendix~\ref{sec:proof}.
Theoretical results show that dynamic adjustment of learning rates, guided by the multimodal balance coefficient, 
consistently accelerates optimization and ensures sufficient learning across modalities.
We also discuss how results can be generalized to various gradient-based optimizers.

\section{Experiment}\label{sec:exp}

\noindent\textbf{Experiment Setup}
We conduct experiments on two non-text modalities, vision and audio, with multiple instruction-tuning downstream tasks: 
(1) for \noindent\textbf{Vision}, we evaluate the backbone MLLMs including BLIP-2 \citep{li2023blip}, InternVL2 \cite{chen2024internvl}, and LLaVA-1.5 \cite{liu2023llava},
on three visual question-answering tasks: TextVQA \citep{singh2019towards}, IconQA \citep{lu2021iconqa}, and A-OKVQA \citep{schwenk2022okvqa}, which focus on text recognition and reasoning, knowledge-intensive QA, 
 and abstract diagram understanding, respectively;
(2) for \noindent\textbf{Audio}, we leverage the SALMONN \citep{tang2023salmonn} model and evaluate one audio question-answering task and two audio captioning tasks: ClothoAQA~\citep{lipping2022clotho}, MACS~\citep{morato2021macs}, and SDD~\citep{manco2023song}, 
which focus respectively on crowdsourced audio question-answering, acoustic scene captioning, and text-to-music generation.
We include detailed implementation details in Appendix \ref{app:implement} for individual backbone MLLM.

We follow the instruction tuning diagram \citep{dai2024instructblip,tang2023salmonn,huang2023vtimellm}, where the parameters of backbone LLMs are finetuned with LoRAs \citep{hu2021lora} and the feature encoders are finetuned directly.
We set the learning rate to $1e-4$ for all the feature encoders and backbone LLMs in our baseline methods \noindent\textbf{Constant LR} \citep{dai2024instructblip,tang2023salmonn}, \noindent\textbf{Feature CD}, \noindent\textbf{Language CD} \citep{wright2015coordinate}.
For Feature CD, we first update the feature encoder until its weights stabilize, then update the backbone LLMs. For Language CD, the process is reversed, with the LLMs being trained first. We also use $1e-4$ as the base learning rates for our CoMMIT variants. There are two CoMMIT variants: \noindent\textbf{CoMMIT} and \noindent\textbf{CoMMIT-CLR}. \noindent\textbf{CoMMIT} is out proposed method in this paper, while \noindent\textbf{CoMMIT-CLR} is an ablation on \noindent\textbf{CoMMIT}, without  the last regularization terms in \eqref{eq:fea-weight} and \eqref{eq:lm-weight}.

\noindent\textbf{LLM Usage}
In this paper, LLMs are only used for refining the writing of natural language.

\begin{figure}[ht]
    \centering
    \includegraphics[width=0.50\textwidth]{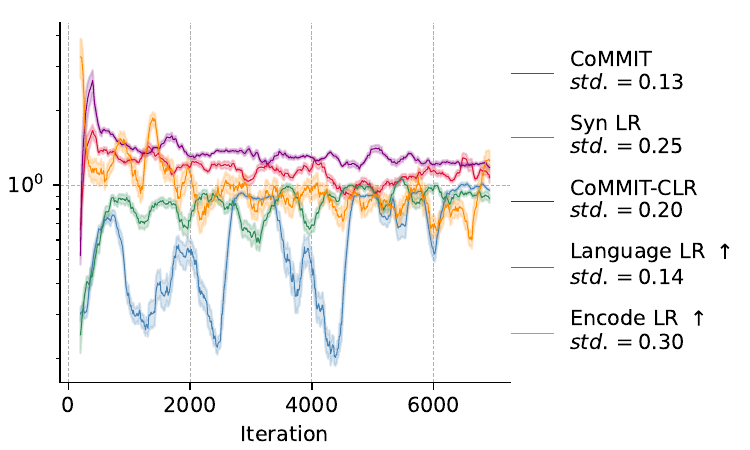}
    \caption{Learning curves of the multimodal learning balance coefficient $\kappa_t$ for multiple methods. In addition to the learning curve, we also report the standard deviation of $\kappa_t$ of each method. }
    \label{fig:exp-kappa}
\end{figure}

\noindent\textbf{Mitigating Oscillation and Biased Learning.}
In Figure \ref{fig:exp-kappa}, we inspect the oscillation problem in Section~\ref{sec:pre-imbalance} by showing the value of $\kappa_t$ with our proposed CoMMIT and CoMMIT-CLR, 
comparing to the three learning rate scheduling methods described in Section \ref{sec:emp}.
We report with the BLIP-2 \citep{li2023blip} backbone model on the task of TextVQA \citep{singh2019towards}.
It can be observed that both CoMMIT and CoMMIT-CLR can stabilize multimodal learning with smaller standard deviations of $\kappa_t$ over time, thus alleviating the oscillation problem in Observation \ref{obs:lr_0} (in Section~\ref{sec:pre-imbalance}).

Though Language LR $\uparrow$ also yields high stability on $\kappa$, 
such learning rate adjustment method suffers from the problems of biased learning as described in Observation \ref{obs:lr_1} (in Section~\ref{sec:pre-insufficient}), 
which potentially causes insufficient training on the feature encoder and worse performance of instruction tuning. 
Comparing CoMMIT and CoMMIT-CLR, we can observe that CoMMIT achieves less biased learning with the value of $\kappa_t$ closer to $1$ 
while demonstrating relatively milder learning oscillation with less variant $\kappa_t$ during training.
Further, we find that the proposed loss regularization in Section \ref{method} also improves the training balance.
Accompanied by the loss regularization and learning balance coefficient $\kappa_t$,
CoMMIT more effectively adapts the optimization process for better training between the feature encoder and LLM.

 \begin{figure*}[ht]
    \centering
    \begin{minipage}{1.\textwidth}
        \centering
        \includegraphics[width=1.\linewidth, page=1]{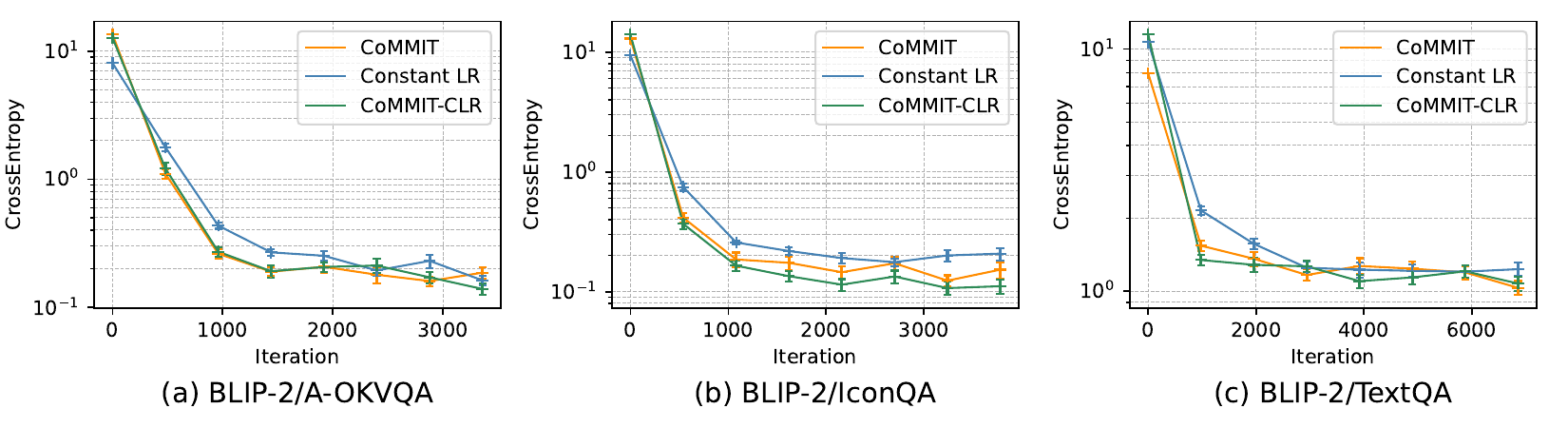}
        \caption{Instruction-tuning learning curves of BLIP-2 on three vision-based downstream tasks.}
        \label{fig:exp-loss-1}
    \end{minipage}
    \hfill
    \vspace{1mm}
    \begin{minipage}{1.\textwidth}
        \centering
        \includegraphics[width=1.\linewidth, page=2]{figures/Loss.pdf}
        \caption{Instruction-tuning learning curves of SALMONN on three audio-based downstream tasks.}
        \label{fig:exp-loss-2}
    \end{minipage}
  \vspace{-3mm}
\end{figure*}

Note that SynLR with the oscillated value of $\kappa_t$ is our baseline Constant LR. In Figure \ref{fig:exp-loss-1} and \ref{fig:exp-loss-2}, 
it is shown that Constant LR generally results in a slower training rate characterized by a flatter descent in loss values during the early stages.
This is consistent with Observation~\ref{obs:lr_0} in Section~\ref{sec:pre-imbalance}, suggesting that the oscillation problem can slow the convergence of MLLMs in instruction tuning.

\noindent\textbf{Improved Convergence in MLLM Instruction Tuning.}
We evaluate the convergence performance of the proposed methods CoMMIT-CLR and CoMMIT in comparison with Constant LR in Figure \ref{fig:exp-loss-1} and \ref{fig:exp-loss-2}.
For visual question-answering tasks in Figure~\ref{fig:exp-loss-1}, we observe that CoMMIT-CLR and CoMMIT can accelerate the instruction tuning of BLIP-2 in the early stage.
This is especially evident in the case of IconQA, which is out-of-domain for BLIP-2's pretraining. Specifically, IconQA requires regional-level and spatial visual understanding, which differs from the tasks BLIP-2 was pre-trained on \citep{chen2023lion}.
In addition, CoMMIT-CLR and CoMMIT achieve lower training losses than Constant LR, validating CoMMIT’s improved training efficiency.
This efficiency gain is attributed to CoMMIT’s ability to mitigate the oscillation problem (Section~\ref{sec:pre-imbalance}), thus accelerating convergence.

Similar to the vision-based tasks, we can find in Figure \ref{fig:exp-loss-2} that CoMMIT-CLR and CoMMIT can also converge to lower loss values in audio tasks.
Specifically, we observe that the CoMMIT-CLR and CoMMIT can achieve better accelerations on the audio captioning tasks of MACS and SDD, compared to training on the audio question-answering task of ClothoAQA.  
Since audio captioning tasks need more adaptation in MLLMs to generate relatively longer context and align the generation distribution with specific tasks,
the coordinated learning rate scheduling method in Section \ref{method} can more dynamically adjust the learning rate for less learned components at each model update step.
In addition, we show that the proposed loss regularization method adopted in CoMMIT can actively promote the difference in MLLM's generation distribution between optimization steps,
which can better benefit tasks, such as audio captioning, that require the model to generate longer contexts. Such improved convergence is associated with our proposed balancing strategy that improves on the oscillation and biased learning problem as discussed.

\begin{table*}[ht]
\centering
\small
\begin{tabular}{l l | c c c | c c}
\toprule
\textbf{Model} &   \textbf{Task}  & \textbf{\makecell{Constant\\LR}} & \textbf{\makecell{Feature\\CD}} & \textbf{\makecell{Language\\CD}} & \textbf{\makecell{CoMMIT\\CLR}} & \textbf{CoMMIT} \\
\midrule
\multirow{3}{*}{\textbf{BLIP-2}} 
& A-OKVQA   & 54.06       & 57.99      & 49.87       & 60.44      & \textbf{64.37} \\
& IconQA    & 37.16       & 35.48      & 34.47       & \textbf{39.09} & 38.65 \\
& TextVQA   & 26.48       & 18.00      & 19.44       & 27.66      & \textbf{28.12} \\
\midrule
\multirow{3}{*}{\textbf{SALMONN}} 
& ClothoAQA & 42.49       & 45.80      & 38.52       & \textbf{52.86} & 50.55 \\
& MACS      & 24.60       & 22.41      & 23.64       & 23.81      & \textbf{25.06} \\
& SDD       & 15.10       & 5.70       & \textbf{15.74} & 15.07      & 15.33 \\
\midrule
\multirow{3}{*}{\textbf{InternVL2}} 
& A-OKVQA   & 76.59 & 73.19 & 79.47 & 78.00 & \textbf{80.52} \\
& IconQA    & 80.94 & \textbf{83.20} & 81.60 & 80.85 & 82.87 \\
& TextVQA   & 65.22 & 65.60 & 65.08 & 65.18 & \textbf{67.00} \\
\midrule
\multirow{3}{*}{\textbf{LLaVA-1.5}} 
& A-OKVQA & 79.20 & 77.64 & 76.94 & 77.82 & \textbf{79.55}	\\
& IconQA  & 64.09 & 64.16 & 58.17 & 65.78  & \textbf{69.60}	 \\
& TextVQA & 41.98 & 43.34 & \textbf{49.32} & 47.80 & 49.30 \\
\bottomrule
\end{tabular}
\caption{
Instruction tuning results for four MLLMs: BLIP-2, SALMONN, InternVL2-8B, and LLaVA-1.5-7B. These are pre-trained LLMs in vision and audio respectively. 
For questions-answering tasks like A-OKVQA, IconQA, TextVQA, and ClothoAQA, we report the accuracy score of the generated answers.
For audio captioning tasks (MACS and SDD), we report the Rouge-L metric that compares the generated caption with candidate captions.
We highlight the best method in bold font for each downstream task of instruction tuning.
}
\label{tab:main}
\vspace{-1em}
\end{table*}

\noindent\textbf{Improved Downstream Performance across Modalities.} In Table \ref{tab:main}, We evaluate the performance of the proposed methods CoMMIT-CLR and CoMMIT, comparing with three baselines Constant LR, Feature CD, and Language CD.
Among the three baselines, we observe that coordinate gradient descend methods have the most improvement compared to the constant learning rate methods that show significant learning tendencies towards a certain modality 
(\emph{e.g.}, Language CD in SDD, and Feature CD in A-OKVQA and ClothoAQA).
However, since such learning balance varies in downstream tasks, coordinate descend methods cannot consistently improve MLLM instruction tuning, 
while arbitrarily inclining towards only a certain modality can result in inferior model performance (\emph{e.g.}, Feature CD in SDD and Language CD in A-OKVQA).

Different from the fixed learning tendency which needs to be predetermined by coordinate descend methods,
the proposed coordinated learning rate scheduling method can dynamically adapt learning rates for multimodal components and balance the multimodal joint training.
With better coordinated multimodal learning, CoMMIT-CLR and CoMMIT consistently improve Constant LR  across modalities and downstream tasks.
In addition, the proposed regularization in CoMMIT can promote larger step sizes in gradient descent, which enlarges differences in the generated output distributions between different time steps.
This prevents learning from being stuck at local optima, which can be especially beneficial for modality-specific captioning tasks whose optimization space can be relatively larger than question-answering tasks.

\begin{table}[t]
\centering
\small
\vspace{-1mm}
\begin{tabular}{l | c c c}
\toprule
\textbf{Method}  & \textbf{A-OKVQA} & \textbf{TextVQA} & \textbf{IconVQA} \\
\midrule
\textbf{LR=1e-5}    & 50.30 & 27.58 & 35.45 \\
\textbf{LR=1e-4} & 54.06 & 26.48 & 37.16 \\
\textbf{LR=1e-3}    & 45.24 & 20.60 & 34.93 \\
\textbf{CoMMIT}     & \textbf{64.37} & \textbf{28.12} & \textbf{38.65} \\
\bottomrule
\end{tabular}
\caption{Comparison on A-OKVQA, TextVQA, and IconVQA with BLIP-2 backbone model. Baselines are Constant LR that direct
fine tune the backbone model with various learning rate. }
\label{tb:lr}
\vspace{-6mm}
\end{table}

\noindent\textbf{Comparing various learning rates.}
In Table \ref{tb:lr}, we compare CoMMIT with results of constant LR with different learning rates. We report the results on tasks of  A-OKVQA, TextVQA, and Icon-VQA with BLIP-2 backbone model.  
We can observe that our proposed CoMMIT outperforms the Constant LR baselines by a significant margin. In addition, we can also find that no fixed value of learning rate consistently yields the best performance for Constant LR, 
while our proposed CoMMIT can dynamically adjust its learning rate. These results demonstrate the necessity and effectiveness of dynamic learning rate adjustment for balanced learning between the feature encoder and LLM in multimodal instruction tuning.

\vspace{-2mm}
\section{Conclusion}
In this work, we address the challenge of imbalanced learning between the feature encoder and the backbone LMM during MLLM instruction tuning. 
Through theoretical analysis and empirical observations, we uncovered how this imbalance can lead to the dilemmas of oscillation and biased learning.
To mitigate these problems, we proposed \textbf{CoMMIT}, a novel approach that dynamically coordinates the learning rates of the feature encoder and LLM backbone.
Our \textbf{CoMMIT} also included regularization on the gradient gradients that promotes training sufficiency.
Our theoretical and empirical analyses demonstrate that \textbf{CoMMIT} improves the balance between the training of the feature encoder and the LLM.
Experiments across multiple vision and audio downstream instruction tuning tasks illustrate that the training with \textbf{CoMMIT} for MLLMs is more effective compared to baselines.

\section{Limitation}

While our framework demonstrates compatibility with open-source LLMs to ensure reproducibility and controlled experimentation, its integration with API-based models remains unexplored.
This focus aligns with common research practices prioritizing transparent benchmarking, though we acknowledge that real-world deployment scenarios may involve complex model ecosystems. 
Future work could investigate adapter mechanisms to bridge this gap, but such extensions lie beyond our current scope of foundational multimodal learning dynamics.

\bibliography{custom}

\appendix
\onecolumn
\section{Theoretical Analysis}\label{sec:theory} 

In this section, we present the computation and proof of a new convergence bound with our proposed method CoMMIT. 
Our theoretical analysis demonstrates that it achieves a faster convergence rate compared to the imbalanced MLLM instruction tuning.

\subsection{Setup and Notations}
Consider a non-convex random objective function $F: \mathbb{R}^d \rightarrow \mathbb{R}$ expressed as the average of $K$ component functions, $F(x) = \frac{1}{K} \sum_{k =1}^K f_k(x)$, where each $f_k(x)$ is an i.i.d sample.
Our goal is to minimize $\esp{F(x)}$ over $x \in \mathbb{R}^d$. 
We also define $\mathbb{E}_{k-1}$ as the conditional expectation with respect to $f_1, f_2, \cdots, f_k$.
Following the notation in Adam \citep{kingma2014adam}, let $m_k$, $v_k$, $x_k \in \mathbb{R}^d$ be vectors at iteration $k$. The $i$-th component of each vector is denoted by $m_{k,i}$, $v_{k,i}$, $x_{k,i}$.
Building upon the insight of D{\'e}fossez et al. \citep{defossez2020simple} regarding the presence of two bias correction terms $m_k$ and $v_k$, we define $\alpha_{k,i} = \alpha_i \sqrt{\frac{1-\beta_2^k}{1- \beta_2}}$. 
Notably, we opt to drop the correction term for $m_k$ due to its faster convergence compared to $v_k$. 

Aligned with our proposed methodology, we incorporate two additional terms into the original Adam algorithm. To prevent learning from oscillating too heavily toward either $S$ or $X$, we adapt $\lambda$ according to the moving average $\Tilde{\kappa}$. 
To mitigate the risk of vanishing or exploding gradients, we introduce $h(x)$ as an extra term in the gradient updates defined in Section~\ref{method} to enhance training stability and support the overall robustness of the learning process.
We fix $\beta_1 = 0$, $ 0 < \beta_2 \leq 1$, $\alpha_{k,i} >0$, $\epsilon = 10^{-8}$, and initialize $m_0 = 0$ and $v_0 = 0$. The updated rules follow,
\begin{align}
    \label{eq:v_update_refine}
    v_{k, i} &= \beta_2 v_{k-1, i} + \left(\lambda \nabla_i f_k(x_{k-1}) +  \nabla_i h_k(x_{k-1}) \right)^2\\
    \label{eq:finalupdate_refine}
    x_{k, i} &= x_{k-1, i} - \alpha_k \frac{\lambda  \nabla_i f_k(x_{k-1}) + \nabla_i h_k(x_{k-1})}{\sqrt{ v_{k, i} + \epsilon}}
\end{align}

Throughout the proof, we assume the norm of the gradients $\|\nabla f(x) + \nabla h(x)\|$ is bounded by $R - \sqrt{\epsilon}$. The small constant $\epsilon$ is used for numerical stability.

\subsection{Necessary Assumptions}\label{asp:bound}
We state the necessary assumptions~\citep{bertsekas2003convex} commonly used when analyzing the convergence of stochastic algorithms for non-convex problems:
\begin{assumption}
The minimum value of $f(x)$ is lower-bounded,
\begin{equation*}
\forall x \in \mathbb{R}^d, \  f^* = \min f(x).
\end{equation*}
\end{assumption}

\begin{assumption}\label{asp:Liptchitz}
The gradient of the non-convex objective function $f$ is $L$-Liptchitz continuous~\citep{nesterov2013introductory}. Then $\forall x, y\in \mathbb{R}^d$, the following inequality holds, 
\begin{equation*}
f(y) \leq f(x) + \nabla f(x)^T (y-x) + \frac{L}{2} \|x-y\|^2_2.
\end{equation*}
\end{assumption}

\subsection{Convergence Analysis}

Following the second Theorem outlined by D{\'e}fossez et al.~\citep{defossez2020simple}, we calculate the convergence bound of our algorithm with a dynamic learning rate and loss function.
\begin{theorem}\label{theom:convergence}
Given the assumptions from Appendix~\ref{asp:bound} and applying Lemma~\ref{sec:descentlemma}, for all components of the step sizes and gradients, update $\alpha_i$ with the corresponding value from $H^S_t$ and $H^T_t$. Let $\{x_k\}$ be a sequence generated by the optimizer, with $ 0 < \beta_2 \leq 1$, and $\alpha_i >0$. For any time step $K$, we have, 
\begin{align}
\esp{\|\nabla F(x_k)_2^2\|} \leq 2R \frac{F(x_0)- f^*}{\lambda \alpha_i K} + C
\end{align}
where $$C = \frac{1}{K} \left(\frac{2 \alpha_i R}{\sqrt{1- \beta_2}} + \frac{\alpha_i^2 L}{2(1 -\beta_2)} \right) \ln \left(\frac{(1-\beta_2^k) R^2}{(1-\beta_2)\epsilon \beta_2} \right) $$
\end{theorem}
The detailed proof is provided in Appendix~\ref{sec:convergence}.

CoMMIT adjusts both $\lambda$ and $h(x)$  to balance multimodal learning progress. The parameter $\lambda$, which measures the balance between feature and language learning, remains greater than 1 during training, driven by the balance metric $\Tilde{\kappa}$. By avoiding learning oscillations, $\lambda$ can grow even larger, contributing to faster learning. When $\Tilde{\kappa}> 1$, the model suffers from insufficient feature learning. CoMMIT reduces the learning rate of the LLM to balance learning, ensuring $\lambda = \frac{\Tilde{\kappa}_t + 1}{\Tilde{\kappa}_{t-1} + 1} > 1$. Conversely, when $\Tilde{\kappa}< 1$, the model suffers from insufficient language learning, ensuring $\lambda = \frac{ 1/ \Tilde{\kappa}_t + 1}{1/ \Tilde{\kappa}_{t-1} + 1} > 1$.
Notably, $\nabla h(x)$ is directly added to $\nabla f(x)$ to induce gradient changes, which further contributes to the increase of $\lambda$, resulting in a faster convergence rate. 

In this section, we show the proof using Adam as the base optimizer. Due to the reason that CoMMIT does not modify the optimization algorithm itself, the theorem can also be extended to any gradient-based optimization method such as the stochastic gradient descent.

\section{Proof}\label{sec:proof}
\subsection{Learning balance in multimodal joint training}\label{sec:lrbalance}
\begin{proof}
According to Lipschitz continuity in cross-entropy loss function \citep{mao2023cross},
there exists a sequence of $T_t$ and $S_t$ during MLLM instruction tuning, where multimodal components are jointly trained.
Given the two metric spaces, $(\mathbb{R},l_2)$ of the cross-entropy losses and $(H,d)$ of the generation distributions, there exists $0 < \gamma < 1$ such that, at each optimization step $t$,
\begin{equation}\label{eq:main}
\left\| L\left(P_{X_t, S_t}\right) - L\left(P_{X_{t+1}, S_{t+1}}\right) \right\|_2 \leq \gamma d\left( P_{X_{t+1}, S_{t+1}} || P_{X_t, S_t}\right),
\end{equation}
where the metric $d$ measures the change in the prediction distribution $P_{X_t, S_t} \in H$ as the multimodal components $X$ and $S$ are updated. 
Based on the triangle inequality in metric space, a joint step of multimodal learning is bounded by the combination of two components' separate step forward,
\begin{equation}\label{eq:sub}
    d\left( P_{X_{t+1}, S_{t+1}} || P_{X_t, S_t} \right) \leq d\left( P_{X_{t+1}, S_t}|| P_{X_t, S_t}\right) + d\left( P_{X_t, S_{t+1}} || P_{X_t, S_t} \right),
\end{equation}
where the first term represents the change due to updating the $X$-component while keeping $S$ fixed, and the second term represents the change due to updating the $S$-component.

Then we can derive the multimodal gradient estimated bounds based on MLLM's generative performance in its metric space $d$ shown in the Theorem~\ref{thm1}.
\end{proof}

\subsection{Controlling Deviation from Descent Direction}\label{sec:descentlemma}

Following the first Lemma outlined by D{\'e}fossez et al.~\citep{defossez2020simple}, where the expected update direction can positively correlate with the gradient~\citep{sashank2018convergence}, we aim to control the deviation from the descent direction to enhance convergence.
\begin{lemma}\label{lemma:descent_la}
For all $k \in \mathbb{N}^*$ and $ R \geq \|\nabla f(x) + \nabla h(x)\| + \sqrt{\epsilon} $, the gradient update follows a descent direction,   
\begin{align}
&\espk{\nabla_i F(x_{k-1}) \frac{ \lambda  \nabla_i f_k(x_{k-1}) + \nabla_i h_k(x_{k-1})}{\sqrt{\epsilon + v_{k,i}}}} - \frac{\lambda (\nabla_i F(x_{k-1}))^2}{2\sqrt{\epsilon + \tilde{v}_{k,i}}} \nonumber\\
&\geq \frac{\nabla_i F(x_{k-1}) \nabla_i h_k(x_{k-1})}{2\sqrt{\epsilon + \tilde{v}_{k,i}}} - 2R \espk{\frac{(\lambda \nabla_i f_k(x_{k-1}) + \nabla_i h_k(x_{k-1}))^2}{\epsilon + v_{k,i}}}.
\end{align}
\end{lemma}

\begin{proof}
Denote $F = \nabla_i F(x_{k-1})$, $f = \lambda  \nabla_i f_k(x_{k-1})$, $h = \nabla_i h_k(x_{k-1})$, and $\tilde{v}_{k,i} = \beta_2 v_{k-1, i} + \espk{\left(\lambda \nabla_i f_k(x_{k-1}) + \nabla_i h_k(x_{k-1}) \right)^2}$, we get:
\begin{equation}\label{eq:descent1}
    \espk{\frac{F(f +h)}{\sqrt{\epsilon + v_{k,i}}}} = \espk{\frac{F(f +h)}{\sqrt{\epsilon + \tilde{v}_{k,i}}}} + \espk{F(f +h)\left(\frac{1}{\sqrt{\epsilon + v_{k,i}}} - \frac{1}{\sqrt{\epsilon + \tilde{v}_{k,i}}}\right)}
\end{equation}
We know that $g$ and $\tilde{v}_{k,i}$ are independent given $f_1, f_2, \cdots , f_{n-1}$. $h$ and $\tilde{v}_{k,i}$ are also independent based on our settings which do not affect the momentum, we have,
\begin{equation}\label{eq:descent2}
\espk{\frac{F(f +h)}{\sqrt{\epsilon + \tilde{v}_{k,i}}}} = \frac{\lambda F^2}{\sqrt{\epsilon + \tilde{v}_{k,i}}} + \frac{Fh}{\sqrt{\epsilon + \tilde{v}_{k,i}}}
\end{equation}
The only thing we need to do is control the deviation of the second term in \eqref{eq:descent1}. Applying Cauchy-Schwarz~\citep{steele2004cauchy},
\begin{align}
    RHS &=  F(f +h) \frac{\espk{(f + h)^2}-(f+ h)^2}{\sqrt{\epsilon + v_{k,i}} \sqrt{\epsilon + \tilde{v}_{k,i}} (\sqrt{\epsilon + v_{k,i}} + \sqrt{\epsilon + \tilde{v}_{k,i}})} \nonumber\\
\label{eq:descent3}
&\leq F(f +h) \frac{\espk{(f + h)^2}}{\sqrt{\epsilon + v_{k,i}}(\epsilon + \tilde{v}_{k,i})} + F(f +h)\frac{(f + h )^2}{\sqrt{\epsilon + v_{k,i}}(\epsilon + \tilde{v}_{k,i})}.
\end{align}
By applying the inequality  $ a b \leq  \frac{1}{2\lambda} b^2+\frac{\lambda }{2}a^2 $ with $\lambda = \frac{\sqrt{\epsilon + \tilde{v}_{k,i}}}{2}$, a = $\frac{F}{\sqrt{\epsilon + \tilde{v}_{k,i}}}$, and $b= \frac{(f + h) \espk{(f + h)^2}}{\sqrt{\epsilon + \tilde{v}_{k,i}}\sqrt{\epsilon + v_{k,i}}}$, the conditional expectation of the first term in \eqref{eq:descent3} can be bounded as,
\begin{align}
\espk{F(f +h) \frac{\espk{(f + h)^2}}{\sqrt{\epsilon + v_{k,i}}(\epsilon + \tilde{v}_{k,i})} } &\leq \espk{ \frac{F^2}{4\sqrt{\epsilon + \tilde{v}_{k,i}}} + \frac{(f + h)^2 \espk{(f + h)^2}^2}{\sqrt{\epsilon + \tilde{v}_{k,i}}^3 (\epsilon + v_{k,i})}} \nonumber\\
& \leq  \frac{F^2}{4\sqrt{\epsilon + \tilde{v}_{k,i}}} + \espk{\frac{(f + h)^2 \espk{(f + h)^2}}{\sqrt{\epsilon + \tilde{v}_{k,i}}(\epsilon + v_{k,i})}} \nonumber\\
\label{eq:descent4}
&\leq  \frac{F^2}{4\sqrt{\epsilon + \tilde{v}_{k,i}}} + R \espk{\frac{(f + h)^2}{\epsilon + v_{k,i}}},
\end{align}
with respect to the fact that $\epsilon + \tilde{v}_{k,i} \geq \espk{(f + h)^2}$ and $\espk{(f + h)^2} \leq R$.

Similarly, applying the inequality $ab \leq \frac{\lambda }{2}a^2  + \frac{1}{2\lambda} b^2$ with $\lambda = \frac{\sqrt{\epsilon + \tilde{v}_{k,i}}}{2 \espk{(f + h)^2}}$, $a = \frac{F(f+h)}{\sqrt{\epsilon + \tilde{v}_{k,i}}}$, and $b = \frac{(f+h)^2}{\epsilon + v_{k,i}}$, the conditional expectation of the second term in \eqref{eq:descent3} can be bounded as,
\begin{align}
\espk{F\frac{(f + h)^2(f +h)}{\sqrt{\epsilon + v_{k,i}}(\epsilon + \tilde{v}_{k,i})}} &\leq \espk{\frac{F^2}{4\sqrt{\epsilon + \tilde{v}_{k,i}}} \frac{(f+h)^2}{\espk{(f + h)^2}} + \frac{\espk{(f + h)^2}}{\sqrt{\epsilon + \tilde{v}_{k,i}}} \frac{(f+h)^4}{(\epsilon + v_{k,i})^2}} \nonumber\\
&\leq \frac{F^2}{4\sqrt{\epsilon + \tilde{v}_{k,i}}} + \espk{\frac{\espk{(f + h)^2}}{\sqrt{\epsilon + \tilde{v}_{k,i}}} \frac{(f+h)^2}{(\epsilon + v_{k,i})}} \nonumber\\
\label{eq:descent5}
&\leq \frac{F^2}{4\sqrt{\epsilon + \tilde{v}_{k,i}}} + R \espk{\frac{(f+h)^2}{\epsilon + v_{k,i}}},
\end{align}
given again $\espk{(f + h)^2} \leq R$.

Putting inequalities \eqref{eq:descent4} and \eqref{eq:descent5} back into \eqref{eq:descent3} gives,
\begin{equation}\label{eq:descent6}
     \espk{F(f +h)\left(\frac{1}{\sqrt{\epsilon + v_{k,i}}} - \frac{1}{\sqrt{\epsilon + \tilde{v}_{k,i}}}\right)} \leq \frac{F^2}{2\sqrt{\epsilon + \tilde{v}_{k,i}}} + 2R \espk{\frac{(f+h)^2}{\epsilon + v_{k,i}}}
\end{equation}
And, therefore, adding \eqref{eq:descent6} and \eqref{eq:descent2} into \eqref{eq:descent1} finishes the proof.
\end{proof}

\subsection{Proof of Convergence}\label{sec:convergence}
In this section, we prove the theorem~\ref{theom:convergence}.
\begin{proof}
Given $\alpha_k = \alpha \sqrt{\frac{1-\beta_2^k}{1- \beta_2}}$, we apply the Assumption~\ref{asp:Liptchitz} and get,
\begin{equation}\label{eq:cg1}
F(x_k)  \leq F(x_{k-1}) - \alpha_k \nabla F(x_{k-1})^T u_k + \frac{\alpha_k^2 L}{2} \|u_k\|^2_2.        
\end{equation}
Since we define the bound $R \geq \|\nabla f(x) + \nabla h(x)\| + \sqrt{\epsilon}$, it follows that $\sqrt{\epsilon + \tilde{v}_{k,i}} \leq R \sqrt{\sum_{j=0}^{n-1} \beta_2^j}$. By applying this inequality, we obtain,
\begin{align}\label{eq:cg2}
&\alpha_k \left(\frac{(\lambda  \nabla_i F(x_{k-1}))^2}{2\sqrt{\epsilon + \tilde{v}_{k,i}}} 
+ \frac{\nabla_i F(x_{k-1}) \nabla_i h_k(x_{k-1})}{2\sqrt{\epsilon + \tilde{v}_{k,i}}}\right) \nonumber\\
&\geq \alpha \left( \frac{(\lambda  \nabla_i F(x_{k-1}))^2}{2R} 
+ \frac{\nabla_i F(x_{k-1}) \nabla_i h_k(x_{k-1})}{2R} \right).        
\end{align}
By taking the conditional expectation, we apply \eqref{eq:cg2} to Lemma~\ref{lemma:descent_la} to derive results from \eqref{eq:cg1},
\begin{align}\label{eq:cg3}
\espk{F(x_{K})} &\leq \espk{F(x_{k-1})} 
-  \frac{\alpha \lambda}{2 R} \|\nabla F(x_k)_2^2\|  \nonumber\\
&-  \frac{\alpha}{2 R} (\nabla F(x_k)^T \nabla h(x_k)) + \left(2 \alpha_k R + \frac{\alpha_k^2 L}{2}\right) \esp{\|u_k\|^2_2}      
\end{align}
Summing the previous inequality over all $k$ and taking the full expectation with respect to the fact that $\alpha \geq \alpha_k \sqrt{1-\beta_2}$. By applying Lemma 5.2 from D{\'e}fossez et al.~\citep{defossez2020simple}, we get the final bound,
\begin{align}
\esp{\|\nabla F(x_k)_2^2\|} \leq 2R \frac{F(x_0)- f^*}{\alpha(1+\lambda)K} + \left(\frac{2 \alpha R}{\sqrt{1- \beta_2}} + \frac{\alpha^2 L}{2(1 -
        \beta_2)} \right) \left(\frac{1}{K} \ln \left(\frac{(1-\beta_2^n) R^2}{(1-\beta_2)\epsilon} \right) - ln(\beta_2)\right)
\end{align}
\end{proof}

\subsection{Implementation Details} \label{app:implement}
We include specific implementation details for each backbone MLLM we used:
\begin{itemize}
    \item \textbf{BLIP-2} consists of a vision Q-Former and a backbone OPT-2.7B LLM \citep{zhang2022opt}. We fine-tune the Q-Former module as the feature encoder and freeze the visual encoder. 
    We apply LoRA to the BLIP-2 model using a rank of 16, an alpha of 32, and a dropout rate of 0.05.
    \item \textbf{InternVL2} consists of the InternViT model as the visual encoder and a backbone InternLM-7B-Chat \citep{chen2024internvl}. We fine-tune the cross-attention layer as the feature encoder, which encodes the encoded visual representations into multimodal tokens, and freeze the visual encoder. We apply LoRA to the BLIP-2 model using a rank of 16, an alpha of 32, and a dropout rate of 0.05.
    \item \textbf{LLaVA-1.5} consists of the CLIP-L-336px as the visual encoder and a backbone Vicuna-7B \citep{chiang2023vicuna}. We fine-tune the projection layer as the feature encoder, which is a linear layer connecting the encoded image and LLM's word embedding space, and freeze the visual encoder. We apply LoRA to the BLIP-2 model using a rank of 16, an alpha of 32, and a dropout rate of 0.05.
    \item \textbf{SALMONN} extracts both speech and audio features from waveforms and composes these low-level features by a learnable audio Q-Former structure as the feature encoder.
    The audio tokens generated by the audio Q-Former are prefixed to the language instruction tokens, which are then input to the backbone Vicuna-7B LLM \citep{chiang2023vicuna}.
\end{itemize}

\textbf{Computation Resources}
Our model is trained on 4 A100 GPUs with 40GB memory. The average training time is about 8 hours.

\end{document}